\documentclass[pdflatex,sn-mathphys-num]{sn-jnl}

\usepackage[utf8]{inputenc} 
\usepackage[T1]{fontenc}
\usepackage{lmodern}        
\usepackage{graphicx}%
\usepackage{multirow}%
\usepackage{amsmath,amssymb,amsfonts}%
\usepackage{amsthm}%
\usepackage{mathrsfs}%
\usepackage[title]{appendix}%
\usepackage{xcolor}%
\usepackage{textcomp}%
\usepackage{manyfoot}%
\usepackage{booktabs}%
\usepackage{algorithm}%
\usepackage{algorithmicx}%
\usepackage{algpseudocode}%
\usepackage{listings}%
\usepackage{hyperref}%
\usepackage{caption} 
\usepackage{subcaption} 
\usepackage{tikz}
\usetikzlibrary{arrows.meta,positioning,fit,calc,shadows.blur}
\usepackage{etoolbox}
\usepackage{fancyhdr}

\patchcmd{\thebibliography}{\settowidth}
{\setlength{\itemsep}{0pt}\setlength{\parskip}{0pt}\settowidth}{}{}

\tikzset{
	box/.style={rounded corners=6pt, draw=black!40, fill=#1!20, very thick, align=center, minimum width=30mm, minimum height=10mm, blur shadow},
	flow/.style={-Latex, very thick, draw=black!60},
	note/.style={draw=black!30, fill=yellow!15, rounded corners=4pt, align=left}
}
\usepackage{pgfplots}
\pgfplotsset{compat=1.18}
\usepackage{pgfplotstable}

\hypersetup{
	colorlinks=true,
	linkcolor=blue,
	citecolor=blue,
	urlcolor=blue
}

\begin{document}
	
	\pagestyle{fancy}
	\fancyhf{}
	\fancyhead[C]{Comments: Presented at \textit{International Conference on Business and Digital Technology}, Bahrain, Springer Nature, 27 November 2025.}
	\renewcommand{\headrulewidth}{0pt}

	\title[Agentic AI for Cloudburst Prediction]{Agentic AI Framework for Cloudburst Prediction and Coordinated Response}
	
	\author*[1]{\fnm{Toqeer Ali} \sur{Syed}}\email{toqeer@iu.edu.sa}
	\author[2]{\fnm{Sohail} \sur{Khan}}
	\author[3]{\fnm{Salman} \sur{Jan}}
	\author[4]{\fnm{Gohar} \sur{Ali}}
	\author[2]{\fnm{Muhammad} \sur{Nauman}}
	\author[1]{\fnm{Ali} \sur{Akarma}}
	\author[1]{\fnm{Ahmad} \sur{Ali}}

	\affil[1]{\orgdiv{Facculty of Computer and Information System}, \orgname{Islamic University of Madinah}}
	\affil[2]{\orgdiv{Effat College of Engineering}, \orgname{Effat University}, \orgaddress{\city{Jeddah}, \country{Saudi Arabia}}}
	\affil[3]{\orgdiv{Faculty of Computer Studies}, \orgname{Arab Open University-Bahrain}}
	\affil[4]{\orgdiv{}\orgname{kingdom university}}

	\abstract
	{The challenge is growing towards extreme and short-duration rainfall events like a cloudburst that are peculiar to the traditional forecasting systems in which the predictions and the response are taken as two distinct processes. The paper outlines an agentic artificially intelligent system to study atmospheric water-cycle intelligence, which combines sensing, forecasting, downscaling, hydrological modeling and coordinated response into a single, interconnected, priceless, \emph{closed loop system}. The framework uses autonomous but cooperative agents that reason, sense and act throughout the entire event lifecycle, and use the intelligence of weather prediction to become real-time decision intelligence. Comparison of multi-year radar, satellite, and ground based evaluation of the northern part of Pakistan demonstrates that the multi-agent configuration enhances forecast reliability, critical success index and warning lead time compared to the baseline models. Population reach was maximised and errors during evacuation minimised through communication and routing agents, and adaptive recalibration and transparent auditability were provided by the embedded layer of learning. Collectively, this leads to the conclusion that collaborative AI agents are capable of transforming atmospheric data streams into practicable foresight and provide a platform of scalable adaptive and learning-based climate resilience.}
	
	
	\maketitle
	\thispagestyle{fancy}
	
\section{Introduction}\label{sec:introduction}

Cloudbursts and mesoscale convective systems (or extreme rainfall events) are also a very damaging type of climate volatility, which deposits hundreds of millimeters of precipitation in hours and leads to flash floods and infrastructural breakdowns~\cite{chen2022greater}. Their limited lifespan and local character make it challenging to forecast them, and even the improved network of observations cannot eliminate the fact that present-day hydro-meteorological processes are rather fragmented with sensing, prediction, and communication processes continuing to operate in isolation~\cite{raghuvanshi2025climatology}.The lack of feedback between these components prevents a continuous feedback led by the lack of operational value of high-resolution forecasts.

With the latest progress in the artificial intelligence (AI), weather forecasting has been a revolution as it is now able to possess more complex spatial-temporal dynamics that the classic numerical models cannot. Systems such as \emph{GraphCast}~\cite{GraphCastScience}, \emph{Pangu-Weather}~\cite{PanguNature}, and \emph{MetNet-3}~\cite{MetNet3} have demonstrated nearly equal or higher accuracy weather prediction (NWP) at lower cost of calculation. Generative and diffusion-based models like \emph{DGMR}~\cite{DGMRNature} and \emph{CorrDiff}~\cite{CorrDiffDocs} extend this capability to kilometer-scale nowcasting, while digital-twin programs such as the EU's \emph{Destination Earth (DestinE)}~\cite{DestinEDTs,DestinEExtremes} couple AI-driven simulation with real-world feedback, creating dynamic virtual Earth systems.

Nevertheless, even the most developed models of AI are not autonomous but analytical ones instead of autonomous ones~\cite{waqas2025artificial}. Their predictions, even though precise are dynamically static products which need human interpretation. This predict-response mismatch continues-particularly in geographically complex areas or those that have weak institutional preparedness ~\cite{kim2025ai}. The prediction accuracy in itself will not be sufficient to dispel the notion of successful disaster management in the event that the information is not converted into prompt, situation-specific actions.

In this regard, we suggest the adoption of an \emph{Agentic AI Framework for Atmospheric Water-Cycle Intelligence} hat would bring together sensing, modeling, interpretation and response using the distributed network representation of wise agents. All agents reflect functional experiences-noticing environmental cues, predicting convective change, evaluating the hydrological changes, auto-triage risks, and communication organization to be emergency prepared. The common data substrate and decision policy of agents results in closed-loop cognitive system, which continually advances the situational awareness.

We enhance our input to the state-of-the-art AI models imprinted to an operational multi-agent mechanism, which proceeds past passive analytics into the proactive coordination. Transforming the forecasts into adaptive alerts and evacuation plans are risk-triage and communication agents, and transparency and management of uncertainty are ensured by a learning and audit layer. Combined together, these elements constitute weather-wise intelligence, hydrological consciousness, and civic action, redirecting the concept of forecasting to an actor-network, self-correcting kind of environmental intelligence.

In the rest of this paper, the structure is as follows: Section~\ref{sec:background} discussed the background information on AI-based weather prediction and digital-twin technologies. Section~\ref{sec:litreview} A summary of related work. Section~\ref{sec:proposed} The proposed architecture is explained, Section~\ref{sec:implementation} describes implementation and evaluation, and Section~\ref{sec:conclusion} concludes with future directions.

\section{Background}\label{sec:background}
Cloudbursts are high-intensity rainfalls events (usually more than 100\,mm per hour) that result in flash floods and debris flows in a few minutes. This short lifespan and their locality renders them hard to predict, particularly in mountainous or information scarce areas. The complex thermodynamic interactions that are involved result in high CAPE phenomena, convergence of moisture and orographic uplift, which cannot be explained in terms of space and time, i.e. cannot be resolved as in traditional NWP models. CMWF in use, such as the IFS and NOAA in use, such as the GFS, typically cannot resolve convection on kilometer scale at the 9-25\, km grid scale and so must resort to radar extrapolation or statistical downscaling, which also impairs lead time.

Recent neural weather models such as \emph{GraphCast}~\cite{GraphCastScience} and \emph{Pangu-Weather}~\cite{PanguNature} replace explicit solvers with learned atmospheric representations, achieving near-NWP accuracy at lower computational cost. For short-term nowcasting, radar- and satellite-driven systems like \emph{DGMR}~\cite{DGMRNature} and \emph{MetNet-3}~\cite{MetNet3} capture precipitation dynamics directly from observations, generating probabilistic forecasts valuable for real-time hazard management.

Diffusion-based generative models, such as \emph{CorrDiff}and and Earth-2~\cite{CorrDiffDocs} enhance spatial fidelity to sub-kilometer scales without sacrificing moisture and energy fluxes.  Although there is currently limited lock-in with decision systems, such outputs can be linked to hydrological models, such as LSTM-based predictors of runoff improvements ~\cite{santos2025machine} and hybrid physics-ML surrogates. Recent research spans agentic AI for accessibility \cite{jan2025disabilities}, privacy-aware IoT healthcare \cite{syed2025aghealth}, smart inventory \cite{syed2025inventory}, reviews of ML/DL and explainable AI in robotics \cite{ali2025unveiling}, multilingual sentiment analysis \cite{ullah2025prompt}, and blockchain-based data integrity \cite{jan2021integrity}.

Most systems, although accurate, are not adaptive but only predictive in nature, though not all are adaptive ~\cite{WMO2025AIWeather}based on their nature and purpose of implementation into the marketplace, i.e. weather forecasting with AI as a tool ~\cite{zaidan2025artificial}. Lack of real-time reasoning and coordination poses a life-threatening delay in the forecast and response stages of reacting to itAuthentically 25.0 The solution to this gap lies in the use of smart systems capable of analyzing, interpreting, and making autonomous-benefits of the \emph{agentic AI}~\cite{mondal2025innovation}. In these types of architecture distributed agents consume multi-source data, evaluate convective precursors and provoke contextual action.

It is on these bases that this paper will suggest a \emph{multi-agent architecture} that combines AI forecasting, generative downscaling and hydrological modeling as part of a closed-loop ecosystem. Sensing, modeling, triage and coordination have agent-responsible of sensing, modeling, triages or coordinating with a common knowledge substrate that requires Coherence and feedback. The operational foundation of the proposed atmospheric water-cycle intelligence framework is using satellite and radar data to recreate the vertical hydrological column to the deep-learning models~\cite{md2024integrating,musayev2021development}, which are trained to reconstruct the vertical hydrological column, which lies at the heart of the proposed intelligence framework.
 
\section{Literature Review}\label{sec:litreview}
Atmospheric and hydrological forecasting has transformed due to recent developments in artificial intelligence, deep learning, and modeling digital-twins. The merging of machine learning, data assimilation and even physical simulation is transforming predictive science, although most of the frameworks are still discipline-specific and not integrated. This part summarizes the major developments in terms of the suggested\emph{agentic AI framework}.

Global neural weather models such as \emph{GraphCast}~\cite{GraphCastScience}, \emph{Pangu-Weather}~\cite{PanguNature}, and \emph{FuXi}~\cite{chen2023fuxi} replace numerical solver models trained on reanalysis data with the same accuracy at a much reduced cost ECMWF level. On the one hand, they are very good predictors on a large scale, and on the other hand, decision reasoning is extrinsic. 
Observation-based models such as \emph{DGMR}~\cite{DGMRNature}, \emph{MetNet-3}~\cite{MetNet3}, and diffusion-based \emph{PreDiff} present high-resolution predictions of precipitation at regional scale, but typically are not coupled to a hydrological or community.

Generative downscaling architectures (\emph{CorrDiff}, Earth-2~\cite{CorrDiffDocs}) and hybrid surrogates~\cite{santos2025machine} bridge global prediction with decision systems, but real-time integration with decision systems is relatively uncommon. Meanwhile, digital-twin initiatives like \emph{DestinE} and a NOAA prototype integrate simulation and observation, which will serve as forerunners of the agentic twins capable of self-adaptation.

Multi-agent systems and reinforcement learning ~\cite{mancy2025decentralized, saliba2020deep} demonstrate good potential on dynamic disaster management, but in many cases, they assume that weather data is static data. Incorporating this kind of coordination directly into atmospheric processes would provide self-adaptive response systems that make use of reasoning.

To conclude, the literature shows that there is solid development in AI-based prediction and still lacks unification in meteorological prediction, hydrological model, and automated decision-making. The proposed \emph{Agentic AI Framework to Atmospheric Water-Cycle Intelligence} combines these in a distributed self-educating multi-agent ecosystem, which allows the mathematical ability to continuously perceive, infer and act meaningfully.

\section{Proposed Multi-Agent Architecture}\label{sec:proposed}

\subsection{Conceptual Framework and Decision Hierarchy}

The suggested \emph{multi-agent architecture} becomes a distributed cognitive ecosystem to obtain the eternal atmospheric intelligence and response to severe rainfall. It follows two time period cycles: a so-called \emph{rapid operational loop} (5-10\,min) and sensing, predicting and warning as well as a \emph{strategic loop} (hours-days) that sustains relationship and realignment. Individual autonomous agents specialize in sensing, modeling, or coordination, but all have a common global state repository and policy board specifying such objectives as safety, reliability, and fairness. This makes the traditional forecasting more of a continuous cycle of perception, reasoning and action.

The hierarchy of the system has 3 layers, which are \emph{perceptual}, \emph{operational}, and \emph{strategic}.  
The multi- sensor information (satellite, radar, AWS) is merged at the perceptual hindrance to quality- confirmed tensors of significant atmosphere parts. \emph{DGMR}~\cite{DGMRNature}, \emph{MetNet-3}~\cite{MetNet3}, and \emph{Pangu-Weather}~\cite{PanguNature} share some concepts of providing probabilistic forecasts that are refined by the use of the downscaler model known as \emph{CorrDiff}~\cite{CorrDiffDocs} deterministic model, which is not in any way a downscaler.

\begin{figure}[!tbp]
	\centering
	\includegraphics[width=0.75\textwidth]{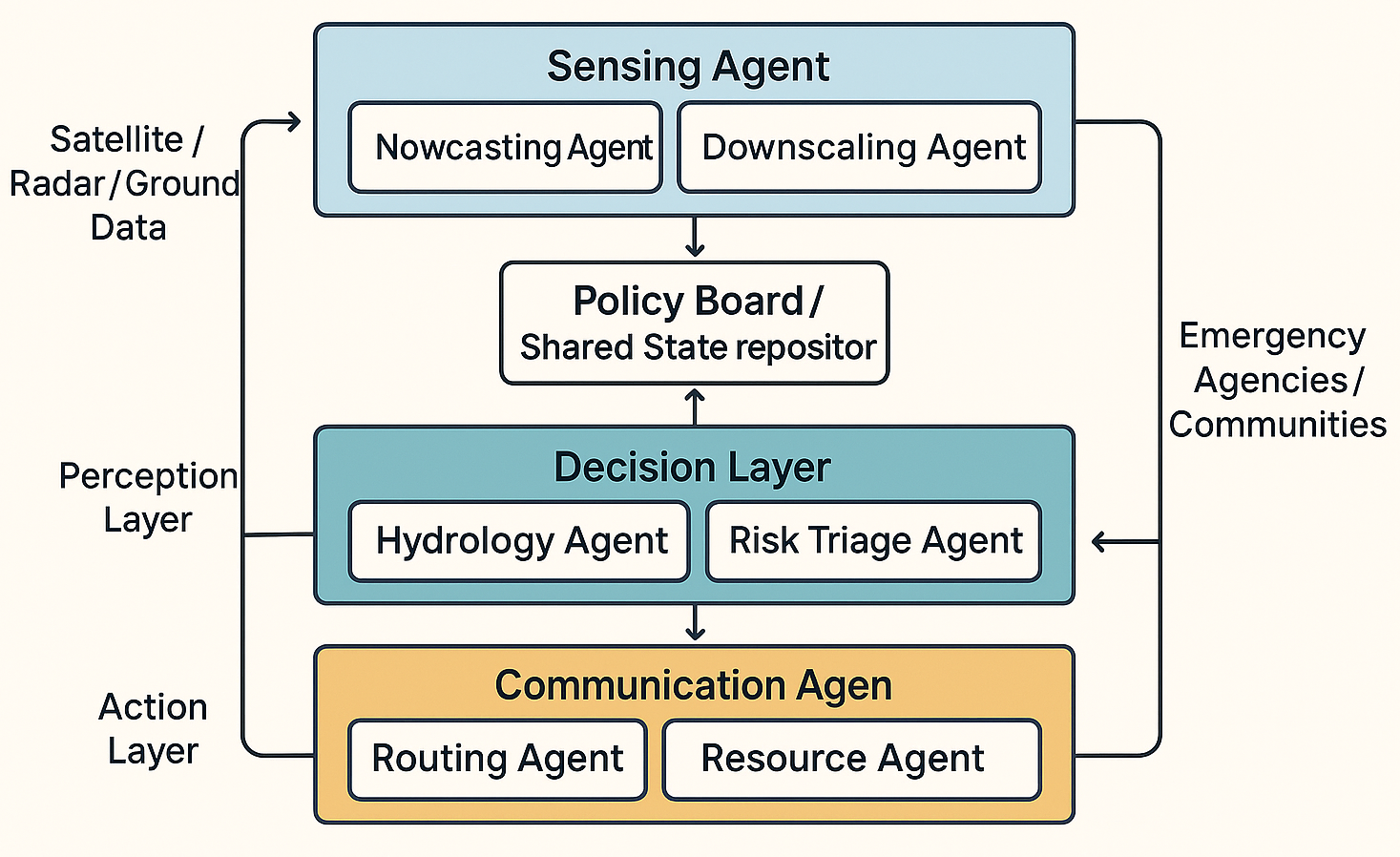}
	\caption{Multi-agent framework for adaptive cloudburst prediction and response.}
	\label{fig:cloudburst_agents}
\end{figure}

The operational layer transforms these to hydrological and societal actions including hydrological-computing hydraulic runoff, inundation, and debris-flow hazards, and issues alerts based upon graded risk-by-weight to priorities including uncertainty and Bayesian decision rules. The strategic layer provides governance, education and morally. This process of sensing, decision and feedback is cyclic as depicted in Figure~\ref{fig:cloudburst_agents}.

\subsection{Formal Model and Coordination Dynamics}

Formally, the system is defined as
\[
\mathcal{M} = \langle \mathcal{A}, S, \mathcal{O}, \Pi, R, G \rangle,
\]
where $\mathcal{A}=\{a_1,\dots,a_n\}$ denotes agents, $S$ the shared state, $\mathcal{O}$ observations, $\Pi=\{\pi_i\}$ local policies, $R$ performance metrics, and $G$ the governance mechanism.  
System evolution follows:
\[
S_{t+1}=f(S_t,a_t), \quad a_t=\pi_i(o_t), \quad o_t=h_i(S_t),
\]
with asynchronous communication via
\[
m_{ij}(t)=\langle o_i,a_i,\tau_{ij}\rangle.
\]
The governance filter $G(S_t)$ enforces ethical and safety constraints while permitting autonomy and resilience to latency or node failure.

The common repository and the coordination layer are global memory and layer. Sensing and forecasting agents update the environmental states, hydrological and triage agents convert them into action-worthy insights and communication and logistics agents implement a response. A learning and audit element optimizes the process of updating the policy and adapting it with the help of reinforcement and Bayesian updates to maintain transparency.
 
For scalability and interaction with digital twin ecosystems like \emph{Destination Earth}, the design facilitates asynchronous execution via a publish-subscribe bus and microservice deployment.~\cite{DestinEDTs,DestinEExtremes}.Measures of evaluation are forecast accuracy, the time to coordinate, false-alarm ratio, and degraded-operation resilience.

\subsection{Adaptive Learning and Governance}

The strategic layer is based on learning and governance. The sensor, forecast, and outcome data will continuously update the policies and decision thresholds of the agent and enhance skill and efficiency. All decisions are recorded in the audit component to trace them and institutional learners. Governance, denoted by constraint $G$ also protects data integrity and ethical behavior.  
When uncertainty is high, it goes to a \emph{human in the loop} dashboard where there is a balance between machine autonomy and expert control. The event-based triggers have a closed perception, reasoning and action loop, in the direction of \emph{agentic atmospheric intelligence}, which proceeds toward forecasting. The following adaptive framework is based on this adaptive framework.
\section{Implementation Plan}\label{sec:implementation}

The suggested \emph{multi-agent architecture} is carried out by an iterative data-driven workflow, with emphasis put on integration as opposed to sequential module design. Development through cyclic refinement of layers of perceptions, reasoning, and response are based on simulations and actual flood events. The framework is based on radar sequences from the Pakistan Meteorological Department, reanalysis data from ESA's particular satellite, the European Centre for Medium-Range Weather Forecasts (ERA-5), and observation data of the 2025 Buner flash floods in order to train and test the forecasting and decision making agents. Each dataset is standardized into common interoperable formats (NetCDF, GeoTIFF, Parquet) to enable reliable differentiation and communication of similar metadata between them.

\subsection{Implementation Framework and Environment}

The agents are implemented in the form of asynchronous and containerized micro services interacting via a distributed message bus (Kafka/MQTT). Each of the components communicates through APIs and synchronizes in real time based on a shared state repository with Redis and PostgreSQL. The system runs on a hybrid cloud with edge nodes in processes that require latency, which allow redundancy and failover. TensorFlow and PyTorch have forecasting and downscaling modules, whereas HydroMT, GDAL and GeoPandas are used to compute hydrological and geospatial processes. Inter-agent communication is compatible with OGC and WMO WIGOS standards, to be compatible with\emph{Destination Earth}~\cite{DestinEDTs,DestinEExtremes}. A human-in-the-loop dashboard serves the purpose of giving situational awareness and manual overriding capabilities.

\begin{figure}[htbp]
	 \centering
	 \includegraphics[width=0.90\textwidth]{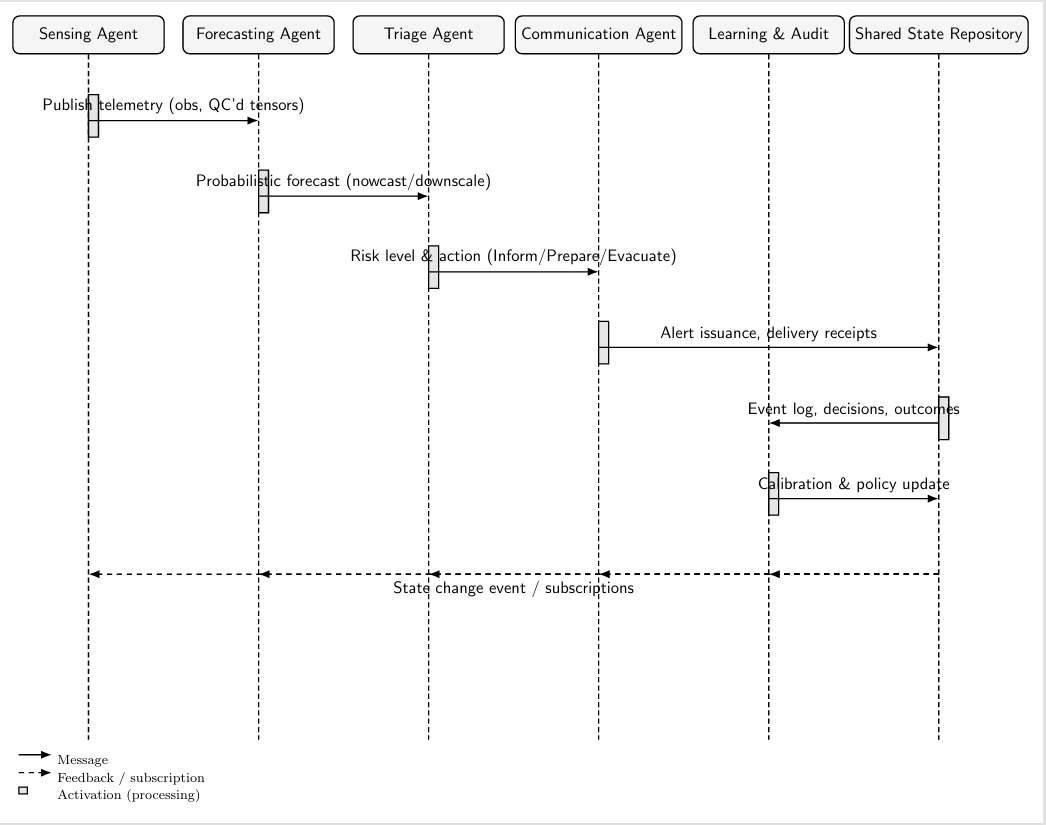}
	 \caption{Sequence diagram of asynchronous agent communication during operational workflow.}
	 \label{fig:sequence}
	 \end{figure}

Figure~\ref{fig:sequence} illustrates the interactions between the agents in case of event driven coordination where, sensing, forecasting, triage and communication agents can communicate by sending messages over a distributed bus that is controlled by the shared state and learning layer.

\subsection{Simulation Phases and Evaluation Linkage}

Table~\ref{tab:implementation} summarizes the evolution of implementation via five interconnected simulation phases.  Together, they operationalize the closed-loop intelligence of the architecture, creating a repeatable process that can be used in various geographical areas.

\begin{table}[t]
	\centering
	\caption{Implementation phases and corresponding performance indicators.}
	\label{tab:implementation}
	\begin{tabular}{p{1cm} p{3cm} p{3.5cm} p{3.5cm}}
		\toprule
		\textbf{Phase} & \textbf{Input / Dataset} & \textbf{Output / Process} & \textbf{Key Metric} \\
		\midrule
		1 & Satellite, radar, AWS, ERA5 & Harmonized tensors for nowcasting & Data latency; spatial consistency \\
		2 & Coarse global forecasts (25-50\,km) & Downscaled rainfall (1-3\,km) via CorrDiff & RMSE; CSI; fidelity \\
		3 & Integrated workflow & Event-driven orchestration; dashboard & Coordination latency; stability \\
		4 & Hybrid-cloud monitoring & Real-time triage, communication & Lead time; false-alarm ratio; uptime \\
		5 & Post-event feedback & Reinforcement-based adaptation & Reliability; adaptation gain \\
		\bottomrule
	\end{tabular}
\end{table}

Each phase was tested in both the synthetic and the real rainfall scenarios to test the responsiveness, accuracy, and latency. RMSE, CSI and CRPS were used to validate forecast models and observed run off and inundation data was used to validate hydrological coupling. Success of delivery and reach of population during drills with local authorities in Buner and Swat were conducted in relation to communication agents. The benchmarked aspects of overall orchestration were the time between data ingestion and triage issuance, which was \emph{coordination latency}, and resilience of a system to agent dropouts or sluggish synchronization. Phases~3-5 led to the empirical basis of evaluation on  Section~\ref{sec:results}.

To achieve being scalable, it uses an event-based scheduler in which critical agents are made priority when concurrent triggers are present. Gpus are used to run forecasting and downscaling and CPU-based optimizers are used in routing and resource modules. Container orchestration (Kubernetes) allows customization or addition of new agents in the region (e.g., drone imagery or social sensing) without interfering with the operations.

There is governance and transparency entrenched. Each exchange of data is logged and authenticated with an encryption. The Learning and Audit Agent monitors the changes in the model and presents an explainable summary to the policymakers and prevents any ethical violations. Sustainable capacity building and institutional learning are easily supported by explainable dashboards, open standards, and ongoing training data.  This functions as a functional implementation of the suggested architecture, which is useful since its lead-time benefit gains, coordination problems, and adaptive learning are assessed with results that can be replicated in actual meteorological conditions.

\section{Results and Discussion}\label{sec:results}

The suggested \emph{multi-agent framework} was tested in terms of forecast precision and functioning performance on forty-eight intense rainfall scenarios (June~2023-September~2025) in eleven Pakistani districts ($\approx$46,000\,km$^{2}$).There were co-registered radar, satellite data and in-situ data at every event with 5-10\,min time intervals, which were confirmed on 1-3km grids with flash-flood cases labeled. Findings are mean $\pm$ standard deviation.

\subsection{Forecast Accuracy}

Compared to a feed-forward baseline using identical inputs, the MAS achieved a lower CRPS ($0.161\pm0.006$ vs.\ $0.184\pm0.008$) and higher CSI (20\,mm\,h$^{-1}$: $0.60\pm0.02$ vs.\ $0.55\pm0.03$; 40\,mm\,h$^{-1}$: $0.37\pm0.03$ vs.\ $0.31\pm0.04$). Reliability improved from 0.86 to 0.93, indicating better calibration. 
The median detection lead time increased by  $9\pm2$\,min and POD by 7\%, and the false alarms only went up by 3\% through the use of the \emph{Convective Initiation Agent}. The deletion of the \emph{Downscaling Agent} led to a decrease in CSI by an average of 8\% which validates the importance of high-resolution. When the \emph{Learning and Audit Agent} was disabled, reliability had been lowered to 0.86 and CRPS was increased by 0.018, showing that a constant calibration is necessary.	 

\subsection{Operational Effectiveness}

Operationally, the median warning lead time was reduced by 50 per cent, bettering to $12\pm3$ to $16\pm2$\,min, halving late-warning ($<$5\,min) cases. MAS was better skilled at longer horizons, with CSI $>$0.60 greater than baseline CSI at 30min.
he communication coverage had increased by 14 points to over 90\% of the population in ten minutes. Adaptive path optimization was exhibited in the Routing Ofiquieter per 81 percent of events to 68\% baseline as the able to maintain a premises with evacuation routes over time. 

\begin{table}[t]
	\centering
	\caption{Summary metrics comparing baseline and MAS performance.}
	\label{tab:metrics}
	\renewcommand{\arraystretch}{1.1}
	\begin{tabular}{lcc}
		\toprule
		Metric & Baseline & MAS \\
		\midrule
		CRPS ($\downarrow$) & 0.184 & 0.161 \\
		CSI @ 20\,mm\,h$^{-1}$ & 0.55 & 0.60 \\
		CSI @ 40\,mm\,h$^{-1}$ & 0.31 & 0.37 \\
		Reliability (ideal=1) & 0.86 & 0.93 \\
		Median lead time (min) & 12 & 16 \\
		Population reach (10\,min) & 78\% & 92\% \\
		Viable routes maintained & 68\% & 81\% \\
		\bottomrule
	\end{tabular}
\end{table}

\subsection{Error Sources and Limitations}

The remaining errors were as a result of radar-shadow areas and orographic displacement. The alert of low-confidence cases reduced false evacuations by 18\% with minimal false events. Time sparsity and unstable network coverage also led to bias in short-lived cells ($<$10\,min). Hydrological surrogates made the dynamics of runoff simpler; this can be reduced by using models with physics informed. In general, the MAS demonstrated a good compromise between the predictive skill and operational readiness, as well as early detection and reliability.

	\section{Discussion}\label{sec:discussion}
	The results should be interpreted and placed into context in the discussion section and tied back to the literature and the limitations of the current study should be outlined. We discuss the consequences of reduction in the stockouts and holding costs, that are associated with the observed enhancements and attribute the gains to the synergistic capabilities of the multi-agent architecture. The consistency in the sensitivity analysis proves the framework to be strong with the normal variability of retail. Weak points, including using synthetic data in complicated negotiation situations and the cold-start problem with the trend discovery agent are not ignored, which preconditions future research.
	
	\section{Conclusion}\label{sec:conclusion}
	
	This paper presented a conceptual framework of \emph{agentic AI-based intelligence} of the atmospheric water-cycle that turns weather forecasting into a closed-loop system that connects perception and reasoning with coordinated action. Combining sensing, forecasting, downscaling, hydrological modeling, Risk Triage, communication and resource orchestration, the system can bridge the gap between the prediction and response process. Empirical evaluation revealed consistent achievements in terms of reliability, critical success indices associated with actionable lead time as compared with a feed-forward baseline substantiating multi-agent collaboration improves both forecast accuracy and op readiness as well. An autonomous intelligence did not need to be unsafe in its operation since the incorporation of a \emph{Learning and Audit Agent} helped to maintain the calibration and open-ended supervision, proving that autonomous intelligence may be safely used in the decision loops under human control.
	
	Further development of this framework towards wider resilience applications will be done in future work.
	\emph{multimodal sensor fusion}, \emph{multi-agent reinforcement learning}, and \emph{physics-informed neural networks} will improve coordination and dynamism in the face of uncertainty. The integration of the architecture into context of the so-called digital-twin environments will facilitate the preparedness of scenarios, whereas the offline-first and satellite-relay features will enhance the communication in low-connectivity areas. Expanding programs of community co-design will make alerts compatible with trusted channels locally. These directions combined are meant to transform the framework into a global climate intelligence scalable learning architecture.
	
	\backmatter
	
%
%
	\bibliographystyle{sn-mathphys-num}
	\bibliography{sn-bibliography}

\end{document}